\documentclass[10pt,twocolumn,letterpaper]{article}

\usepackage{iccv}
\usepackage{times}
\usepackage{epsfig}
\usepackage{graphicx}
\usepackage{amsmath}
\usepackage{amssymb}

\usepackage{csquotes}

\usepackage{multirow}
\usepackage{graphicx}
\usepackage{booktabs}
\usepackage{bbm}
\usepackage{subcaption,amsfonts,dcolumn}

\usepackage{color, colortbl}

\DeclareMathOperator*{\argmax}{argmax}

\usepackage{array}
\usepackage{tabularx}

\usepackage{algorithm}
\usepackage{algorithmic}
\usepackage{multirow}
\usepackage{xcolor} 

\DeclareMathOperator*{\argmin}{arg\,min}

\usepackage{color, colortbl}
\definecolor{Gray}{gray}{0.9}

\usepackage{pifont}
\newcommand{\cmark}{\ding{51}}%
\newcommand{\xmark}{\ding{55}}%

\usepackage[accsupp]{axessibility}  

\usepackage[pagebackref=true,breaklinks=true,colorlinks,bookmarks=false]{hyperref}

\iccvfinalcopy 

\def\iccvPaperID{6577} 

\ificcvfinal\fi

\begin{document}

\title{Black-box Unsupervised Domain Adaptation with Bi-directional Atkinson-Shiffrin Memory}

\author{Jingyi Zhang \ \  Jiaxing Huang \ \ Xueying Jiang \ \  Shijian Lu\thanks{Corresponding author} 
\\
S-lab, Nanyang Technological University
\\
{\tt\small \{Jingyi.Zhang, Jiaxing.Huang, Shijian.Lu\}@ntu.edu.sg, xueying003@e.ntu.edu.sg}
}

\maketitle
\ificcvfinal\thispagestyle{empty}\fi

\begin{abstract}
Black-box unsupervised domain adaptation (UDA) learns with source predictions of target data without accessing either source data or source models during training, and it has clear superiority in data privacy and flexibility in target network selection. However, the source predictions of target data are often noisy and training with them is prone to learning collapses. We propose BiMem, a bi-directional memorization mechanism that learns to remember useful and representative information to correct noisy pseudo labels on the fly, leading to robust black-box UDA that can generalize across different visual recognition tasks. BiMem constructs three types of memory, including sensory memory, short-term memory, and long-term memory, which interact in a bi-directional manner for comprehensive and robust memorization of learnt features. It includes a forward memorization flow that identifies and stores useful features and a backward calibration flow that rectifies features' pseudo labels progressively. Extensive experiments show that BiMem achieves superior domain adaptation performance consistently across various visual recognition tasks such as image classification, semantic segmentation and object detection.
Codes will be released at \href{https://github.com/jingyi0000/BiMem}{https://github.com/jingyi0000/BiMem}.
\end{abstract}

\section{Introduction}\label{introduction}

Unsupervised domain adaptation (UDA) has been studied extensively in recent years, aiming to alleviate data collection and annotation constraint in deep network training~\cite{chen2018wild,vu2019advent,inoue2018weakly,saito2019strong,xu2020category,2020coarse2fine,ganin2015grl,saito2018maximum,pinheiro2018unsupervised,saito2019semi,zou2019confidence}. However, most existing UDA methods could impair data privacy and confidentiality~\cite{liang2022dine} as they require to access source data or source-trained models during training. In addition, most existing UDA imposes the same network architecture ($i.e.$, the source model architecture) in adaptation which limits the flexibility of selecting different target networks in UDA~\cite{liang2022dine}. 
Black-box UDA only requires the initial predictions of target data provided by black-box source models~\cite{liang2022dine} ($i.e.$, only the model predictions of target data are available~\cite{liang2022dine}) for domain adaptation, as shown in Table~\ref{table:intro}.
It has attracted increasing attention in recent years~\cite{liang2022dine,liang2021source} due to its advantages in data privacy and flexibility of allowing different target networks regardless of the source-trained black-box models.

\begin{figure}[t]
\centering
\includegraphics[width=1.\linewidth]{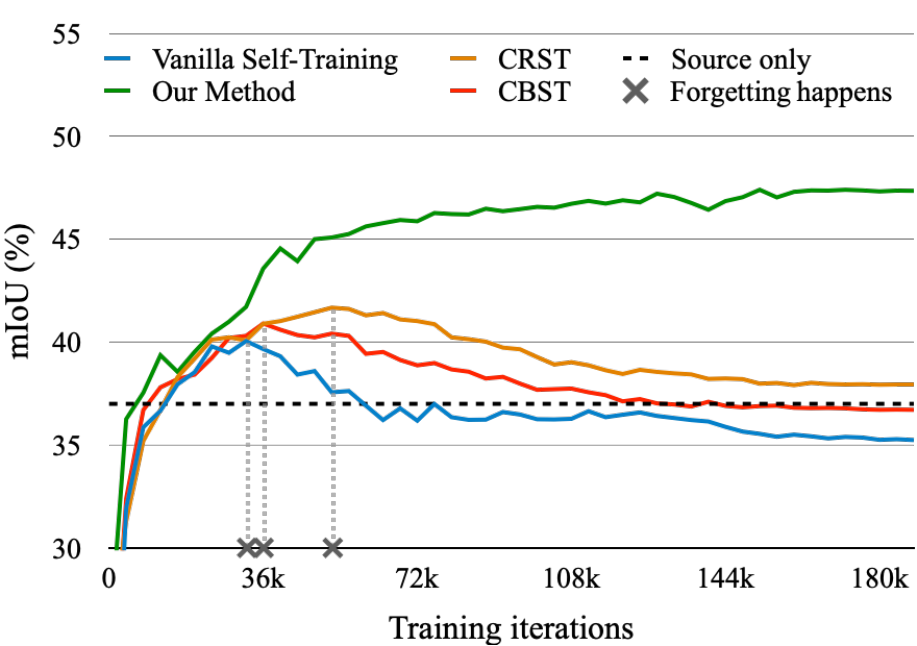}
\caption{
Self-training including Vanilla self-training~\cite{lee2013pseudo}, and advanced CBST~\cite{zou2018self_seg} and CRST~\cite{zou2019confidence} tends to `forget’ learnt useful features due to the accumulation of noisy pseudo labels along the training process in black-box UDA 
– it learns well at the early adaptation stage but collapses with the adaptation moving on and the finally adapted models in~\cite{lee2013pseudo,zou2018self_seg} cannot even compare with the \textit{Source only}. 
We introduce bi-directional memory to remember useful and representative features learnt during adaptation, which 
helps calibrate noisy pseudo labels on the fly and leads to stabler black-box UDA without collapse. The experiments were conducted over domain adaptive semantic segmentation task GTA5 $\rightarrow$ Cityscapes and the evaluations were performed over the Cityscapes validation data.
}
\label{fig:forget}
\end{figure}

\begin{table*}[t]
\centering
\begin{footnotesize}
\begin{tabular}{c|cccccc}
 \toprule
 \multicolumn{6}{c}{Comparisons of Different UDA Setups} \\
 \midrule
Adaptation Setups & Source Data & Source-Trained Model & Source-Predicted Target Labels & Target Data & Privacy Risk \\
 \midrule
 Conventional UDA & \cmark & \cmark & \cmark & \cmark & High  \\
 \midrule
Source-free UDA & \xmark & \cmark & \cmark & \cmark & Medium\\
\midrule
\rowcolor{Gray}
\textbf{Black-box UDA} & \xmark & \xmark & \cmark & \cmark & Low \\

\bottomrule
\end{tabular}
\end{footnotesize}
\caption{
Comparison of different UDA setups: Black-box UDA better preserves data privacy, requiring neither source data nor source-trained models but just source-predicted labels of target data during domain adaptation. It also allows different target networks regardless of source networks.
}
\label{table:intro}
\end{table*}

However, the black-box predictions of the target data are prone to errors due to the cross-domain discrepancy, leading to a fair amount of false pseudo labels. 
As a result, self-training~\cite{lee2013pseudo,zou2018self_seg,zou2019confidence} with the pseudo-labelled target data is often susceptible to collapse as illustrated in Fig.~\ref{fig:forget}.
We argue that the learning collapse is largely attributed to certain `forgetting’ in network training. Specifically, the self-training with noisy pseudo labels learns well at the early stage and the trained model 
outperforms the \textit{Source only} over the test data as shown in Fig.~\ref{fig:forget}. However, the performance deteriorates gradually and drops even lower than that of the \textit{Source only} as the training moves on.
This shows that the self-training learns useful target information and adapts well towards the target data at the early training stage, but then gradually forgets the learnt useful target knowledge and collapses with the accumulation of pseudo-label noises along the training process.

\begin{figure*}[ht]
\centering
\includegraphics[width=0.94\linewidth]{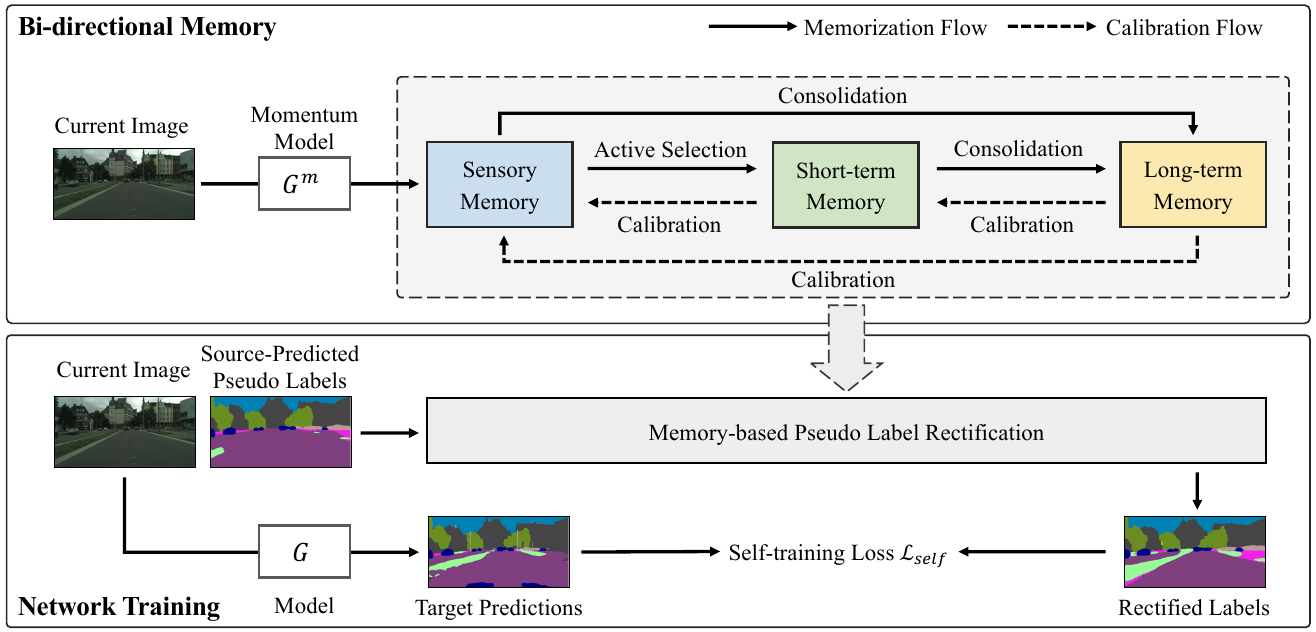}
\caption{
Overview of the proposed BiMem.
BiMem constructs three types of memory that interact in a bi-directional manner via a forward memorization flow and a backward calibration flow as shown in the top part.
In the forward memorization flow, input images are fed to momentum model $G^m$ and the generated features are exploited to construct three types of memory for comprehensive
memorization, capturing fresh yet representative information during the adaptation.
In the backward calibration flow, the pseudo labels of memorized features are calibrated progressively for robust memorization.
Thus, BiMem builds comprehensive and robust memory that allows to rectify source-predicted pseudo labels conditioned on the features stored in BiMem as shown in the bottom part, leading to stable Black-box UDA and better adapted target models.
}
\label{fig:architecture}
\end{figure*}

Inspired by Atkinson-Shiffrin memory~\cite{atkinson1968human}, we propose BiMem, a bi-directional memorization mechanism that constructs three types of memories to address the `forgetting’ problem in black-box UDA. The three types of memory interact in a bi-directional manner including a forward memorization flow and a backward calibration flow as illustrated in Fig.~\ref{fig:architecture}. In the forward memorization flow, the short-term memory actively identifies and stores hard samples ($i.e.$, samples with high prediction uncertainty) from the sensory memory which buffers fresh features from the current training batch. Meanwhile, the long-term memory accumulates features from the sensory and short-term memory, leading to comprehensive memorization that captures fresh yet representative features. In backward calibration flow, we calibrate the pseudo labels of memorized features progressively where the short-term memory is calibrated by the long-term memory while the sensory memory is corrected by both short-term and long-term memory, leading to robust memorization via a progressive calibration process. Hence, BiMem builds comprehensive and robust memory that allows to learn with more accurate pseudo labels and produce better adapted target model as illustrated in Fig.~\ref{fig:forget}.

The contributions of this work can be summarized in three aspects. \textit{First},
we design BiMem, a general black-box UDA framework that works well on different visual recognition tasks. To the best of our knowledge, this is the first work that explores and benchmarks black-box UDA over different visual recognition tasks.
\textit{Second}, we design three types of memory that interact in a bi-directional manner, leading to less `forgetting’ of useful and representative features, more accurate pseudo labeling of target data on the fly, and better adaptation in black-box UDA. \textit{Third}, extensive experiments over multiple benchmarks show that BiMem achieves superior performance consistently across different computer vision tasks including image classification, semantic segmentation, and object detection.

\section{Related Work}
\label{related_works}

\textbf{Unsupervised Domain Adaptation (UDA)} has been extensively studied in various visual recognition tasks for mitigating the data annotation constraint in deep network training ~\cite{ganin2015grl,chen2018wild, zou2018self_seg,saito2019strong,he2019MAF,vu2019advent,yang2020fda,xu2020category,zhuang2020ifan,2020coarse2fine,li2020SAP,huang2020contextual,melas2021pixmatch,zhang2021rpn,zhang2021proda,huang2021rda,araslanov2021self,mirza2022norm,zhang2022spectral,chen2022reusing,huang2022category,gao2022cross,li2022class,hoyer2022daformer,xie2022towards,ye2022unsupervised,yang2023tvt,yu2022sc,zhang2023detr,huang2021fsdr,zhang2023unidaformer}. 
Conventional UDA requires to access the source data in training, which may not be valid while facing concerns in data privacy and confidentiality. Source-free UDA~\cite{liang2020shot,huang2021model,liu2021source,fleuret2021uncertainty, li2020model,xia2021adaptive,yang2021generalized,yeh2021sofa,kundu2020universal,tian2021vdm,ding2022source,wang2022exploring,yang2021generalized,kurmi2021domain,lee2022confidence} addresses the data privacy concerns by adopting a source-trained model instead of source data in domain adaptation, but it still requires source models from which source data could be recovered via certain generation techniques~\cite{goodfellow2014generative}. Beyond data privacy, most conventional and source-free UDA imposes the same network architecture in domain adaptation which precludes the flexibility of selecting different target networks in UDA. We study black-box UDA, a new UDA setup that only requires initial predictions of target data during domain adaptation, hence having little concern of data privacy but great flexibility in target network selection.

\textbf{Black-box UDA} learns with source predictions of target data without requiring either source data or source models during domain adaptation~\cite{liang2022dine}. It has recently attracted increasing attention as it has little data privacy concerns and allows flexible selection of target networks. Most existing black-box UDA studies focus on image classification tasks~\cite{lipton2018detecting,liang2021source,liang2022dine}. For example, \cite{liang2021source} splits target data into two parts according to the prediction confidence and handles unconfident data by adopting a semi-supervised learning strategy. \cite{liang2022dine} distills knowledge via adaptive label smoothing and fine-tunes the distilled model to fit the target distribution.
Beyond image classification, ATP~\cite{wang2021source} introduces negative pseudo labels into self-training and proposes information propagation for tackling black-box UDA for semantic segmentation.
Different from these studies~\cite{liang2021source,liang2022dine}, BiMem focuses on addressing the intrinsic `forgetting' issue in black-box UDA and it is generic and achieves superior performance across various visual recognition tasks.

\textbf{Memory-based Learning} has been widely explored in computer vision~\cite{li2017learning,weston2014memory,kirkpatrick2017overcoming,liu2021rmm,he2020momentum,huang2021model,laine2016temporal,tarvainen2017mean,liu2022learning,wang2020cross,santoro2016meta,zhu2018compound,kim2021robust,park2020learning,belouadah2019il2m,wang2019memory,ko2021learning,hu2021learning}. 
For the task of UDA, several recent studies employ memory to facilitate cross-domain adaptation
~\cite{huang2021model,vs2021mega,kalluri2022memsac}.
For example, \cite{huang2021model} tackles the source-free UDA challenge by memorizing historical models which helps 
the target model to benefit from the source-domain knowledge.
\cite{vs2021mega} stores historical data to generate category-specific attention maps for learning discriminative target representations, which effectively facilitates unsupervised domain adaptation for object detection. 
Different from these studies~\cite{huang2021model,vs2021mega,he2020momentum}, BiMem relies on neither source data nor source models but constructs three types of memory using only target data, which interact with each other in a bi-directional manner for black-box UDA. It aims to build comprehensive yet robust memorization to mitigate the intrinsic `forgetting’ issue in black-box UDA. 

\section{Method}

\subsection{Preliminaries of Atkinson-Shiffrin Memory}

Human memory is powerful in encoding, storing and retrieving information, which plays a fundamental role in human learning. In the field of Neuroscience, Atkinson-Shiffrin memory~\cite{atkinson1968human} models human memory by three types of memory including sensory memory, short-term memory and long-term memory. Among the three, sensory memory stores sensory information ($e.g.$, visual information) which resides for a very brief period of time ($e.g.$, 0.5–1.0 second) and then decays. Short-term memory buffers information selected from sensory memory, where the information in short-term memory has a longer duration ($e.g.$, 18-20 seconds) before being lost. Long-term memory consolidates, compacts and stores information from short-term memory, where the information in long-term memory is relatively permanent. Inspired by the powerful human memory mechanism in human learning, we follow Atkinson-Shiffrin memory theory and build three types of memory for tackling the `forgetting' issue in Black-box UDA. Specifically, we propose a bi-directional memorization mechanism that enables the three types of memory to interact with each other, leading to comprehensive and robust memorization, less `forgetting’ of useful and representative information, and better adaptation in black-box UDA.

\subsection{Task Definition}

This work focuses on black-box UDA for visual recognition tasks including image classification, semantic segmentation, and object detection. 
As shown in Table~\ref{table:intro}, in training, black-box UDA does not access either source data or source-trained model but just pseudo-label predictions of unlabelled target data. It involves a \textit{black-box predictor} that is pre-trained with certain large-scale training data and stored on cloud as a service provider (for pseudo label prediction). In our study, we employ a source-trained model to simulate the black-box predictor which just provides pseudo-label prediction of target data but is not accessible while training the domain adaptation network as in~\cite{liang2022dine}.
We take the single-source scenario for the simplicity of illustrations, which involves a target domain
$\mathcal{D}_t = \{ X_t, \hat{Y}_t\}$, where the pseudo label $\hat{Y}_t$ of target sample $X_t$ is inferred from one black-box source model $G_s$.
The goal of black-box UDA is to train a target model $G$ that well performs on $X_t$. 

\noindent \textbf{Black-box Source Model Generation.} 
We train a source model $G_s$ that will act as a black-box predictor to provide the initial predictions of target data for domain adaptation as in~\cite{liang2022dine}. Generally, the source model $G_s$ is trained with the labelled data in source domain
$\mathcal{D}_s = \{ X_s, Y_s\}$
by a supervised loss:
\begin{equation}
\mathcal{L}_{sup} = l({G_s}(X_{s}), Y_{s}), 
\label{eq:baseline}
\end{equation}
where $l(\cdot)$ denotes a task-related loss, $e.g.$, the standard cross-entropy loss for image classification.

\begin{algorithm}[t]
\footnotesize
    \caption{The proposed BiMem for black-box UDA.}\label{algorithm_BiMem}
    \begin{algorithmic}[1]
        \REQUIRE 
        Target data $X_t$ with source-predicted pseudo labels $\hat{Y}_t$ 
        \ENSURE Learnt target model $G$
        \STATE Initialize $G$ and $G^m$
        \FOR{$iter = 1$ \textbf{to} $Max\_Iter$}
            \STATE Update the momentum model: $G^{m} \leftarrow G$
            \STATE Sample $\{x_t, \hat{y}_t\} \subset \{X_t, \hat{Y}_t\}$ and encode $x_t$: $f_{sm} =G^m(x_t)$
            \STATE \textbf{Forward Memorization Flow:}
            \STATE Update \textit{sensory memory}: replace old features $f_{sm}^*$ 
            by $f_{sm}$
            \STATE Update \textit{short-term memory}: dequeue oldest features $f_{st}^*$ and enqueue new features selected by Eq.~\ref{eq:entropy}
            \STATE Update \textit{long-term memory}: accumulate $f_{sm}^*$ and $f_{st}^*$ by Eqs.~\ref{eq:compact}-\ref{eq:update}
            
            \STATE \textbf{Backward Calibration Flow:}
            \STATE Calibrate \textit{short-term memory} by \textit{long-term memory} as in Eqs.~\ref{eq:5}-\ref{eq:6}
            \STATE  Correct \textit{sensory memory} by \textit{long-term memory} and the calibrated \textit{short-term memory} as in Eq.~\ref{eq:7}
            \STATE \textbf{Network Training:}
            \STATE Acquire $\overline{y}_t$ by denoising $\hat{y}_t$ with \textit{sensory memory} by Eq.~\ref{eq:8}
            \STATE Optimize target model $G$ with $\{x_t, \overline{y}_t\}$ by Eq.~\ref{eq:9}
        \ENDFOR
  \RETURN $G$
  \end{algorithmic}
\end{algorithm}

\subsection{BiMem}

BiMem constructs three types of memory including sensory memory, short-term memory and long-term memory, which interact in a bi-directional manner for mitigating the `forgetting' of useful and representative features during black-box adaptation.
Specifically, \textit{Sensory Memory} stores fresh information along the adaptation by buffering the features of current batch.
\textit{Short-term Memory} actively identifies and stores the hard samples from sensory memory according to samples' uncertainty.
\textit{Long-term Memory} consolidates sensory and short-term memories via category-wise compaction and accumulation of all the features dequeued from sensory memory and short-term memory, which memorizes the global and representative information along the whole adaptation process. 

Taking a batch of $K$ target samples $x_t = \{ x_t^k\}_{k=1}^K$ ($x_t^{k} \in X_t$) and the corresponding source-predicted labels $\hat{y}_t = \{\hat{y}_t^k\}_{k=1}^K$ ($\hat{y}^k_t \in \hat{Y}_t$) as an example. We elaborate the memory construction and update in \textit{Forward Memorization}, and memory calibration in \textit{Backward Calibration}.

\noindent \textbf{Forward Memorization} aims to construct comprehensive memories that capture fresh yet representative information during the adaptation.

We employ a momentum model to encode images for stable memorization as the slow and smooth update mechanism in momentum model allows generating consistent features along the training process.
Specifically, we feed $x_t$ into the momentum model $G^m$ (the moving averaged of $G$, $i.e.$, $\theta_{G^m} \leftarrow \gamma \ \theta_{G^m} + (1 - \gamma) \theta_{G}$, and $\gamma$ is a momentum coefficient) to acquire the momentum features $f_{sm} = \{f^k_{sm} \}_{k=1}^K$ and the corresponding category predictions $\{\{ p^{(k,c)}_{sm} \}_{c=1}^{C}\}_{k=1}^{K}$, where $C$ denotes the number of categories.

Sensory memory buffers $\{f_{sm}, p_{sm}\}$ and it is updated in every iteration, $i.e.$, the features of previous batch of samples $f_{sm}^*$ will be replaced by the features of current batch of samples $f_{sm}$, which allows sensory memory to capture fresh knowledge learnt by the model. 

\textit{Active Selection.} Short-term memory actively selects and stores hard features from the sensory memory. We identify hard samples that are difficult to be classified and generally with high classification uncertainty, $i.e.$, the sample with high uncertainty is considered as a hard sample. Specifically, we measure the uncertainty of $K$ samples by their prediction entropy:
\begin{equation}
\mathcal{H}(f^k_{sm}) = -\sum^C_{c=1} p_{sm}^{(k,c)} \ \log \ p_{sm}^{(k,c)},
\label{eq:entropy}
\end{equation}
where the Top-$N$ samples with the highest entropy ($i.e.$, lowest certainty) are selected and stored in short-term memory.

Short-term memory works as a FIFO queue with a fixed size of $M$ ($M$\textgreater $N$), where $N$ oldest features $f_{st}^*$ will be dequeued 
and the fresh $N$ features selected via Eq.~\ref{eq:entropy} will be enqueued to update the short-term memory $f_{st} = \{ f_{st}^m\}_{m=1}^M$. Note that the short-term feature $f_{st}^m$ is stored with its category prediction $p_{st}^m$. Such FIFO update strategy can avoid GPU memory explosion as the features are uncompressed before storing in short-term memory.

\textit{Consolidation.} Long-term memory stores global and representative information along the whole adaptation process by accumulating all the features dequeued from sensory memory and short-term memory.

Inspired by human memory consolidation mechanism~\cite{squire2015memory}, we consolidate temporary memories ($i.e.$, sensory and short-term memories) into long-term memory for more stable and sustained long-term memorization.
Specifically, we compact the features dequeued from sensory and short-term memories into category-wise feature centroids $\delta_{lt} = \{\delta_{lt}^c\}_{c=1}^{C}$, where the feature centroid $\delta_{lt}^c$ of each category is computed as the following:
\begin{equation}
\delta_{lt}^c = \frac{\sum_{f \in {f_{sm}^* \cup f_{st}^*} }f \cdot \mathbbm{1}(\hat{c} = c)}{\sum_{f \in {f_{sm}^* \cup f_{st}^*}} \mathbbm{1}(\hat{c} = c)},
\label{eq:compact}
\end{equation}
where 
$\mathbbm{1}$ is an indicator function that returns `1' if $f$ belongs to $c$-th category, and `0' otherwise.

To allow the long-term memory to memorize all the information along the adaptation process, we update it with $\delta_{lt}$ in a momentum way:
\begin{equation}
\delta_{lt} \leftarrow  (1-\gamma') \ \delta_{lt} + \gamma{'} \ \delta_{lt}^*,
\label{eq:update}
\end{equation}
where $\delta_{lt}^*$ denotes the old long-term features and $\gamma'$ is a coefficient for smooth feature update in the long-term memory.

With the backward calibration (described in following paragraphs), long-term memory consolidates the calibrated sensory and short-term memories iteratively, the features stored in which tends to gradually move closer to the true feature centroid of each category while the adaptation moves on.
Meanwhile, the consolidation operations specially consider the features of hard samples (with calibrated pseudo labels) that are generally sparse, which allows the long-term features to be more representative.

\noindent \textbf{Backward Calibration} rectifies the memories progressively, aiming to suppress the false pseudo labels predicted for the stored features for robust memorization.
Specifically, we first employ the representative information accumulated in the long-term memory to correct short-term memory. Then, long-term memory and the corrected short-term memory collaborate to calibrate sensory memory.

We calibrate short-term memory by re-weighting the predicted category probability  $p_{st}=\{{p}_{st}^c\}_{c=1}^C$ of short-term features $f_{st}$ as the following:
\begin{equation}
 \overline{p}_{st}^{c} = w^{c} \otimes p_{st}^{c}, 
 \label{eq:5}
\end{equation}
where $\otimes$ denotes the element-wise multiplication and $w^{c} \in \{w^c \}_{c=1}^C $ is the calibration weight for the corresponding $c$-th category probability.
The calibration weight $w$ for each short-term feature $f^m_{st}$ is calculated according to the distance between the short-term feature and the category centroids stored in the long-term memory. Generally, if the short-term feature $f^m_{st}$ is far from the $c$-th long-term feature $\delta^c$, this feature should be assigned
with a low probability of belonging to the $c$-th category. Therefore, the calibration weight in Eq.~\ref{eq:5} is defined as the following:
\begin{equation}
 w^{(m,c)} = \text{Softmax} (-||f^m_{st} - \delta_{lt}^c ||_{1}) , 
 \label{eq:6}
\end{equation}
where $w^c = \{w^{(m,c)}\}_{m=1}^M$ and $||\cdot||_{1}$ denotes $L1$ distance and the softmax operation is performed along the category dimension.

Next, we employ long-term memory and the calibrated short-term memory to correct sensory memory by assigning each sensory feature $f_{sm}^k \in f_{sm}$ a new category probability vector.
Specifically, we first compute the centroids $\delta_{st} = \{\delta^c\}_{c=1}^C$ of hard features over the calibrated short-term memory $\{f_{st}, \overline{p}_{st}\}$ by adopting 
Eq.~\ref{eq:compact}.
With the long-term feature centroids $\delta_{lt}$ and the short-term feature centroids $\delta_{st}$, we assign a new category probability to each sensory  feature $f_{sm}^k$ as the following:
\begin{equation}
 \overline{p}_{sm}^{(k,c)} = \text{Softmax} [ (-||f_{sm}^k - \delta_{lt}^c ||_{1})+(-||f_{sm}^k - \delta_{st}^c ||_{1}) ],
 \label{eq:7}
\end{equation}
where the softmax operation works along the category dimension.

\begin{table*}[t]
\centering
\resizebox{1.0\linewidth}{!}{
\begin{tabular}{c|ccccccccccccccccccc|c}
 \toprule
 \multicolumn{21}{c}{\textbf{GTA5 $\rightarrow$ Cityscapes Semantic Segmentation}} \\
 \midrule
 Methods  & Road & SW & Buil. & Wall & Fence & Pole & TL & TS & Veg. & Ter. & Sky & PR & Rider & Car & Truck & Bus & Train & Mot. & Bike & mIoU\\
 \midrule
 Source only~\cite{he2016resnet} &75.8	&16.8	&77.2	&12.5	&21.0	&25.5	&30.1	&20.1	&81.3	&24.6	&70.3	&53.8	&25.4	&49.9	&17.2	&25.9	&\textbf{6.5}	&25.3	&36.0	&36.6\\
 \midrule
CBST~\cite{zou2018self_seg} &82.7&22.4&70.3&29.1&21.9&21.7&25.9&23.3&76.5&22.3&76.9&55.2&26.4&65.5&36.7&43.3&0.0&26.9&37.1&40.3
\\

CRST~\cite{zou2019confidence} & 87.1 & 48.7& 74.7& 27.4 &16.5 & \textbf{38.5} & 35.8& 23.9 & 84.9 & 36.1 & 67.3 & \textbf{60.9} & 24.4 & 82.1 & 23.1 & 29.3 & 0.4 & 23.7 & 7.4 & 41.7 
\\
SFDA~\cite{liu2021source}& 92.2&51.8&81.0&3.8&23.6&23.2&18.9&25.8&83.9&35.1&84.1&52.3&24.9&81.4&29.5&38.3&0.1&32.8&40.4& 43.3\\
UR~\cite{fleuret2021uncertainty} &92.7&53.8&81.5&10.4&23.2&25.6&16.5&30.8& 84.3& 36.9&83.5&55.6&24.9&82.7& 32.5&40.2&0.5&32.7&44.6 &44.8\\
HCL~\cite{huang2021model}&93.3&58.0&81.9&23.8&24.5&24.9&8.5&31.4&84.2&37.4&84.6&57.4&24.2&84.1& 29.1& 39.9&0.0&33.1&47.5 &45.7\\
TAP~\cite{wang2021source} & 83.6 &25.8& 81.9& 30.2& 25.2& 27.9& \textbf{36.2}& 28.7& 84.8& 34.4& 77.5 &62.2 &35.7 &81.5 &32.3 &16.8& 0.0& 41.7& \textbf{53.5} &45.3 \\
DINE~\cite{liang2022dine}&88.2& 44.2& \textbf{83.5}& 14.1& \textbf{32.4}&23.5&24.6& \textbf{36.8}&\textbf{85.4}&38.3&\textbf{85.3}&59.8&\textbf{27.4}&84.7&30.1&42.2&0.0&\textbf{42.7}&45.3&46.7 \\
\rowcolor{Gray}
\textbf{BiMem}&\textbf{94.2}& \textbf{59.5}&81.7&\textbf{35.2}&22.9&21.6&10.0&34.3&85.2&\textbf{42.4}&85.0&56.8&26.4&\textbf{85.6}&\textbf{37.2}&\textbf{47.4}&0.2&39.9&50.9& \textbf{48.2}\\
\bottomrule
\end{tabular}
}
\caption{Experiments on semantic segmentation over black-box UDA task GTA5 $\rightarrow$ Cityscapes.}
\label{table:gta2city}
\end{table*}

\begin{table*}[t]
\centering
\resizebox{1.0\linewidth}{!}{
\begin{tabular}{c|cccccccccccccccc|c|c}
 \toprule
 \multicolumn{19}{c}{\textbf{SYNTHIA $\rightarrow$ Cityscapes Semantic Segmentation}} \\
 \midrule
 Methods  &{Road} &{SW} &{Buil.}&{{Wall*}} &{Fence*} &{Pole*} &{TL} &{TS} &{Veg.} &{Sky} &{PR} &{Rider} &{Car} &{Bus} &{Mot.} &{Bike} &mIoU  &mIoU*\\
 \midrule
 Source only~\cite{he2016resnet} &55.6	&23.8	&74.6	&\textbf{9.2}	&0.2	&24.4	&6.1	&12.1	&74.8	&79.0	&55.3	&19.1	&39.6	&23.3	&13.7	&25.0	&33.5	&38.6 \\
\midrule
CBST~\cite{zou2018self_seg} &76.7& 30.5 & 69.7 & 8.4 & 0.3 & 31.6 & 0.1 & \textbf{23.2} & 78.4 & 75.7 & 50.1 & 20.1 & 74.1 & 18.6 & 10.0 & 21.3 & 36.8 & 42.1

\\
CRST~\cite{zou2019confidence} &65.9& 26.4& 71.3& 6.7& 0.1 & 33.8 & 10.8 & 24.1 & 81.6 & 79.9 & 53.2 & 15.9 & 74.8 & 19.7 & 12.9 & 21.9 & 37.5& 43.0
\\
SFDA~\cite{liu2021source} &77.2&\textbf{32.7}&74.0&0.7&\textbf{0.4}&34.9&13.9&20.9&82.8&79.1&52.5&\textbf{21.4}&74.7&14.7&11.2&23.1&38.8&44.5\\
UR~\cite{fleuret2021uncertainty} & 74.2&29.0&75.3&0.2&0.0&38.6&17.8&15.0&81.4&53.6&65.9&13.5&74.6&30.0&33.9&\textbf{26.6}&39.5&45.4\\
HCL~\cite{huang2021model}&\textbf{83.8}& 32.5&\textbf{80.7}&0.3& 0.2 &28.3&12.1&5.6&84.1&\textbf{81.4}&60.3&15.0&\textbf{82.9}&25.4&16.3&25.5&40.8&46.6\\
DINE~\cite{liang2022dine} &77.5 & 29.6&79.5&4.3&0.3&39.0&21.3&13.9&81.8&68.9&66.6&13.9&71.7&33.9&\textbf{34.2}&18.6&40.9&47.0\\
\rowcolor{Gray}
\textbf{BiMem}  &78.8 &30.5& 80.4 &5.9&0.1& \textbf{39.2}& \textbf{21.6}& 15.0& \textbf{84.7}& 74.3& \textbf{66.8}& 14.1& 73.3& \textbf{36.0}& 32.3& 21.8& \textbf{42.2} &\textbf{48.4}
\\
\bottomrule
\end{tabular}
}
\caption{Experiments on semantic segmentation over black-box UDA task SYNTHIA $\rightarrow$ Cityscapes. Following previous studies~\cite{zou2018self_seg,zou2019confidence}, mIoU is evaluated on 16 classes while mIoU* is evaluated on 13 classes.}
\label{table:synthia2city}
\end{table*}

\begin{table}[t]
\centering
\resizebox{1.0\linewidth}{!}{
\begin{tabular}{c|cccccccc|c}
 \toprule
 \multicolumn{10}{c}{\textbf{Cityscapes $\rightarrow$ Foggy cityscapes Object Detection}} \\
\midrule
 Methods &pers. & rider & car & truck & bus & train & mot. & bike & mAP\\
\midrule
Source only~\cite{zhu2020deformable}&37.7& 39.1& 44.2& 17.2& 26.8& 5.8 &21.6 &35.5 &28.5\\
\midrule
WSOD~\cite{inoue2018weakly}& 39.9&40.1&50.6&16.5&34.9&8.1&25.2&38.2&31.7
\\
WST~\cite{kim2019self} & 40.7&40.4&53.8&17.1&35.0&5.1&30.2&39.1&32.8
\\
SFOD~\cite{li2021free} & 39.2&39.3&51.8&21.7&33.6&12.5&31.2&\textbf{42.9}&34.0
\\
LODS~\cite{li2022source}&41.5&42.0&54.5&20.5&37.2&23.9&27.1&40.8&35.9
 \\
DINE~\cite{liang2022dine}& 41.4&40.9&55.0&21.8&38.5&25.7&28.2&40.4&36.5

\\
\rowcolor{Gray}
\textbf{BiMem}&\textbf{42.2}&\textbf{42.5}&\textbf{56.9}&\textbf{23.4}&\textbf{39.7}&\textbf{28.5}&\textbf{32.4}&41.3&\textbf{38.4}
\\
\bottomrule
\end{tabular}
}
\caption{Experiments on object detection over black-box UDA task Cityscapes $\rightarrow$ Foggy Cityscapes.}
\label{table:det_city2fog}
\end{table}

\subsection{Network Training}

With the comprehensive and robust memorization built by BiMem, 
we calibrate the source-predicted pseudo labels $\hat{y}_t$ of each sample in $\{x_{t}^k\}_{k=1}^{K}$ as shown in the bottom part of Fig.~\ref{fig:architecture}.
For sample $x_t^k$, we read out its calibrated category-wise probabilities $\overline{p}_{sm}^{(k,c)}$ (acquired by Eq.~\ref{eq:7}) from sensory memory.
Next, the pseudo label $\hat{y}_t^k$ of $x_t^k$ is denoised by re-weighting its category-wise probability $\hat{p}^c$:
\begin{equation}
    \overline{y}_t^k = \argmax_c ( \ \overline{p}_{sm}^{(k,c)} \otimes \hat{p}^{(k,c)} ),
    \label{eq:8}
\end{equation}
where the denoised pseudo labels $\overline{y}_t$ of a batch of $K$ samples $x_t$ can be obtained by rectifying each sample independently.

With the rectified label $\overline{y}_t$, the model $G$ is optimized with the unsupervised self-training loss defined as the following:
\begin{equation}
\label{eq:9}
\mathcal{L}_{self} = l(G(x_t), \overline{y}_t),
\end{equation}
where $l(\cdot)$ denotes a task-related loss, $e.g.$, the standard
cross-entropy loss for image classification. The overall training objective is to minimize the unsupervised self-training loss $\mathcal{L}_{self}$, $i.e.$, $\argmin_G \mathcal{L}_{self}$, as detailed in Algorithm~\ref{algorithm_BiMem}.

\section{Experiment}

This section presents experiments including datasets, implementation details, benchmarking over the tasks of semantic segmentation, object detection and image classification, 
as well as discussion of specific parameters and designs.
More details are to be described in the ensuing subsections.

\renewcommand\arraystretch{.95}
\begin{table}[!t]
\centering
\resizebox{1.0\linewidth}{!}{
\begin{tabular}{c|cccccc|c}
 \toprule
 \multicolumn{8}{c}{\textbf{SYNTHIA $\rightarrow$ Cityscapes Object Detection}} \\
\midrule
 Methods &person & rider & car & bus & mot. & bike & mAP\\
\midrule
Source only~\cite{zhu2020deformable}& 32.7&19.4&34.7&17.2&5.1&21.8&21.8\\
\midrule
WSOD~\cite{inoue2018weakly}& 36.8&21.5&39.3&19.9&4.5&23.2&24.2\\
WST~\cite{kim2019self} & 35.8&21.9&40.0&23.1&3.9&23.4&25.6\\
SFOD~\cite{li2021free} & 39.0&26.5&42.2&24.5&2.9&25.4&26.7\\
LODS~\cite{li2022source}&37.3&30.5&39.6&27.9&6.3&25.8&27.9 \\
DINE~\cite{liang2022dine}&\textbf{41.2}&26.0&45.3&\textbf{28.4}&\textbf{6.8}&26.7&29.1
 \\
\rowcolor{Gray}
\textbf{BiMem} &40.8&\textbf{36.7}&\textbf{49.6}&27.5&5.7&\textbf{28.9}&\textbf{31.5}
\\
\bottomrule
\end{tabular}
}
\caption{Experiments on object detection over black-box UDA task SYNTHIA $\rightarrow$ Cityscapes.}
\label{table:det_syn2city}
\end{table}

\subsection{Datasets}

We evaluate BiMem over multiple datasets across three widely studied computer vision tasks as listed:

\noindent\textbf{Black-box UDA for Semantic Segmentation:} We study two domain adaptive semantic segmentation tasks GTA5 \cite{richter2016playing} $\rightarrow$ Cityscapes \cite{cordts2016cityscapes} and SYNTHIA \cite{ros2016synthia} $\rightarrow$ Cityscapes.

\noindent\textbf{Black-box UDA for Object Detection:}
We study two domain adaptive detection tasks Cityscapes~\cite{cordts2016cityscapes} $\rightarrow$ Foggy Cityscapes~\cite{sakaridis2018foggy} and SYNTHIA~\cite{ros2016synthia} $\rightarrow$ Cityscapes~\cite{cordts2016cityscapes}.

\noindent\textbf{Black-box UDA for Image Classification:} We study two domain adaptive image classification tasks Office-Home~\cite{venkateswara2017deep} and Office-31~\cite{saenko2010adapting}. Office-home consists of 12 adaptation tasks across 4 domains: Art, Clipart, Product and Real-world.
Office-31 includes 6 adaptation tasks across 3 domains: Amazon, DSLR and Webcam.

We provide more details of the involved datasets in the appendix.

\renewcommand\arraystretch{.95}
\begin{table*}[t]
\centering
\resizebox{1.0\linewidth}{!}{
\begin{tabular}{c|cccccccccccc|c}
 \toprule
 \multicolumn{14}{c}{\textbf{Office-Home Classification}} \\
\midrule
 Methods&  Ar $\rightarrow$ Cl& Ar$\rightarrow$ Pr& Ar $\rightarrow$ Re& Cl$\rightarrow$ Ar& Cl$\rightarrow$ Pr& Cl$\rightarrow$ Re & Pr$\rightarrow$ Ar& Pr$\rightarrow$ Cl& Pr$\rightarrow$ Re& Re $\rightarrow$ Ar& Re$\rightarrow$ Cl& Re$\rightarrow$ Pr& Avg.\\
 \midrule
Source only~\cite{he2016resnet}& 44.1&66.9&74.2&54.5&63.3&66.1&52.8&41.2&73.2&66.1&46.7&77.5&60.6\\
\midrule
NLL-OT~\cite{asano2019self} &49.1 &71.7 &77.3 &60.2 &68.7 &73.1 &57.0 &46.5 &76.8 &67.1 &52.3 &79.5 &64.9\\
NLL-KL~\cite{zhang2021unsupervised} &49.0&71.5&77.1&59.0 &68.7& 72.9 &56.4 &46.9 &76.6&66.2 &52.3 &79.1 &64.6\\
HD-SHOT~\cite{liang2020shot}& 48.6& 72.8 &77.0 &60.7& 70.0& 73.2& 56.6& 47.0& 76.7& 67.5& 52.6& 80.2& 65.3\\
SD-SHOT~\cite{liang2020shot} & 50.1 &75.0& 78.8& 63.2& 72.9& 76.4& 60.0& 48.0& 79.4& 69.2& 54.2& 81.6& 67.4\\
DINE~\cite{liang2022dine}&52.2 &78.4&81.3&65.3&76.6&78.7&62.7&49.6&82.2&69.8&55.8&84.2&69.7\\
\rowcolor{Gray}
\textbf{BiMem}  & \textbf{54.5} &\textbf{78.8}& \textbf{81.4}& \textbf{66.7}& \textbf{78.7}&\textbf{79.6}&\textbf{65.9}&\textbf{53.6}&\textbf{82.3}& \textbf{73.6}& \textbf{57.8}& \textbf{84.9}&\textbf{71.5}

\\
\bottomrule
\end{tabular}
}
\caption{Experiments on image classification over black-box UDA task Office-Home.}
\label{table:office_home}
\end{table*}

\subsection{Implementation Details}
\label{implement}

\noindent \textbf{Semantic Segmentation:} We adopt DeepLab-V2~\cite{chen2017deeplab} with ResNet-101 \cite{he2016resnet} as the segmentation network as in~\cite{tsai2018learning,zou2018self_seg}. 

\noindent \textbf{Object Detection:} We adopt deformable-DETR~\cite{zhu2020deformable} with ResNet-50~\cite{he2016resnet} as detection network as in~\cite{cai2019mtor, zhu2020deformable}.

\noindent\textbf{Image Classification:} Following~\cite{liang2022dine}, we adopt ResNet-50~\cite{he2016resnet} for the tasks Office-Home and Office-31.

We provide more implementation details in the appendix.

\subsection{Black-box UDA for Semantic Segmentation}
\label{segmentation}

We evaluate BiMem over two black-box domain adaptive semantic segmentation tasks GTA5 $\rightarrow$ Cityscapes and SYNTHIA $\rightarrow$ Cityscapes.
As there is little prior black-box UDA research on semantic segmentation, we benchmark BiMem by reproducing conventional UDA methods~\cite{zou2018self_seg,zou2019confidence} and source-free UDA methods~\cite{liu2021source,fleuret2021uncertainty,huang2021model}.
In addition, we rerun the SOTA black-box UDA approach~\cite{liang2022dine} (designed for image classification task) on semantic segmentation task for benchmarking.
Tables~\ref{table:gta2city} and~\ref{table:synthia2city} present the experimental results. It can be observed that BiMem achieves superior segmentation performance clearly, largely because BiMem builds comprehensive and robust memorization that helps calibrate the noisy pseudo label and mitigate the `forgetting' issue in Black-box UDA.

\renewcommand\arraystretch{.95}
\begin{table}[t]
\centering
\resizebox{1.0\linewidth}{!}{
\begin{tabular}{c|cccccc|c}
 \toprule
 \multicolumn{8}{c}{\textbf{Office-31 Classification}} \\
 \midrule
 Methods&  A $\rightarrow$ D& A$\rightarrow$ W& D $\rightarrow$ A& D$\rightarrow$ W& W$\rightarrow$ A& W$\rightarrow$ D& Avg.\\
 \midrule
Source only~\cite{he2016resnet}& 79.9&76.6&56.4&92.8&60.9&98.5&77.5\\
\midrule
NLL-OT~\cite{asano2019self} &88.8 &85.5& 64.6& 95.1& 66.7& 98.7& 83.2\\
NLL-KL~\cite{zhang2021unsupervised} &89.4& 86.8& 65.1& 94.8& 67.1& 98.7& 83.6\\
HD-SHOT~\cite{liang2020shot}&86.5 &83.1 &66.1 &95.1 &68.9 &98.1 &83.0\\
SD-SHOT~\cite{liang2020shot}&89.2 &83.7& 67.9 &95.3 &71.1& 97.1& 84.1\\
DINE~\cite{liang2022dine}&91.6&86.8&72.2&96.2&73.3&98.6&86.4
\\
\rowcolor{Gray}
\textbf{BiMem} & \textbf{92.8}&\textbf{88.2}&\textbf{73.9}&\textbf{96.8}&\textbf{75.3}&\textbf{99.4}&\textbf{87.7}
\\
\bottomrule
\end{tabular}
}
\caption{Experiments on image classification over black-box UDA task Office-31.}
\label{table:office31}
\end{table}

\renewcommand\arraystretch{.95}
\begin{table}[t]
\centering
\resizebox{\linewidth}{!}{
\begin{tabular}{c|ccc|ccc|c}
\toprule
Row&\multicolumn{3}{c|}{\textbf{Forward Memorization Flow}}
 &\multicolumn{3}{c|}{\textbf{Backward Calibration Flow}} &\multicolumn{1}{c}{\multirow{2}{*}{mIoU}} 
\\
\cmidrule{2-7}
No.&\multicolumn{1}{c}{SM $\rightarrow$ ST} & \multicolumn{1}{c}{SM $\rightarrow$ LT}& \multicolumn{1}{c|}{ST $\rightarrow$ LT} & \multicolumn{1}{c}{SM $\leftarrow$ ST } & \multicolumn{1}{c}{SM $\leftarrow$ LT} & \multicolumn{1}{c|}{ST $\leftarrow$ LT} &  
\\
\midrule
1&& & & & & & 35.2  \\
\midrule
2&\checkmark & & & \checkmark & & & 44.3\\
3&& \checkmark & & & \checkmark & & 44.5\\
4&\checkmark & \checkmark & & \checkmark & \checkmark & & 46.4 \\
\midrule
5&\checkmark & \checkmark & \checkmark & \checkmark & \checkmark & & 47.5\\
6&\checkmark & \checkmark &  & \checkmark & \checkmark & \checkmark & 47.4 \\
\rowcolor{Gray}
7&\checkmark & \checkmark & \checkmark & \checkmark & \checkmark & \checkmark & \textbf{48.2} \\

\bottomrule
\end{tabular}
}
\caption{
Ablation study of BiMem over GTA5 $\rightarrow$ Cityscapes semantic segmentation task, where `SM', `ST' and `LT' stand for sensory memory, short-term memory and long-term memory, respectively.
}
\label{ablation_1}
\end{table}

\subsection{Black-box UDA for Object Detection}
\label{detection}

We evaluate BiMem over two black-box domain adaptive object detection tasks Cityscapes $\rightarrow$ Foggy Cityscapes and SYNTHIA $\rightarrow$ Cityscapes.
Similar to semantic segmentation benchmarking, we reproduce UDA-based object detection methods ~\cite{inoue2018weakly,kim2019self,li2021free,li2022source} and rerun the SOTA black-box UDA approach~\cite{liang2022dine} 
(designed for image classification task) 
for comparisons.
As shown in Tables~\ref{table:det_city2fog}-\ref{table:det_syn2city}, BiMem achieves superior detection performance as well, indicating the superior generalization of the proposed BiMem.

\subsection{Black-box UDA for Image Classification}
\label{classification}

Following~\cite{liang2022dine}, we evaluate BiMem over two popular black-box UDA classification tasks Office-Home and Office-31. Tables~\ref{table:office_home}-\ref{table:office31} show experimental results. It can be observed that BiMem outperforms state-of-the-art methods clearly.
The superior performance is largely attributed to the memory mechanism in BiMem which helps produce more accurate pseudo labels while learning domain adaptive models.

\subsection{Ablation Studies}

We conduct extensive ablation studies to examine how different BiMem designs contribute to the overall performance. 
Table~\ref{ablation_1} shows experimental results over domain adaptive semantic segmentation task GTA5 $\rightarrow$ Cityscapes. 
We can see that the conventional self-training~\cite{lee2013pseudo} in the 1st Row does not perform well due to the `forgetting' issue in black-box UDA.

We first study how the interaction between sensory memory and short-/long-term memory affects the adaptation performance.
Specifically, on top of the conventional self-training, further including the memorization and calibration flows between sensory memory and either short-term memory or long-term memory improves the performance clearly, as shown in Rows 2-3. This shows that either the hard features captured in short-term memory or the representative features accumulated in long-term memory provides useful information for denoising the initial source-predicted pseudo labels.
Besides, we can see that the network performs clearly better when sensory memory interacts with both short-term memory and long-term memory in Row 4, demonstrating that short-term memory and long-term memory complement each other as they capture different features with complementary information.

Next, we investigate how the memorization and calibration flows between short-term and long-term memories benefit Black-box adaptation.
Specifically, the memorization flow from short-term memory to long-term memory that specially compacts and accumulates the hard features into long-term memory brings clear improvements as shown in Row 5, indicating that the long-term features become more representative as hard features are generally rare. Besides, employing long-term memory to denoise pseudo labels in short-term memory also improves the performance by a large margin as shown in Row 6, showing that hard features in short-term memory are noisy and can be calibrated by long-term features effectively.
At last, BiMem that includes all the memorization and calibration flows performs clearly the best, demonstrating that the proposed bi-directional flow enables comprehensive yet robust memorization which leads to stable and effective Black-box UDA.

\subsection{Discussion}
\label{Discuss}

\noindent \textbf{Generalization across Different Tasks:} Our BiMem is general and works well over various computer vision tasks consistently without any task-specific modifications and fine-tuning as shown in Sections~\ref{segmentation}-\ref{classification}.
The superior generalization capability of BiMem is largely attributed to its task-agnostic underlying mechanism and designs that enable BiMem to work consistently across different tasks.

\begin{figure}
  \centering

    \includegraphics[width=1.0\linewidth]{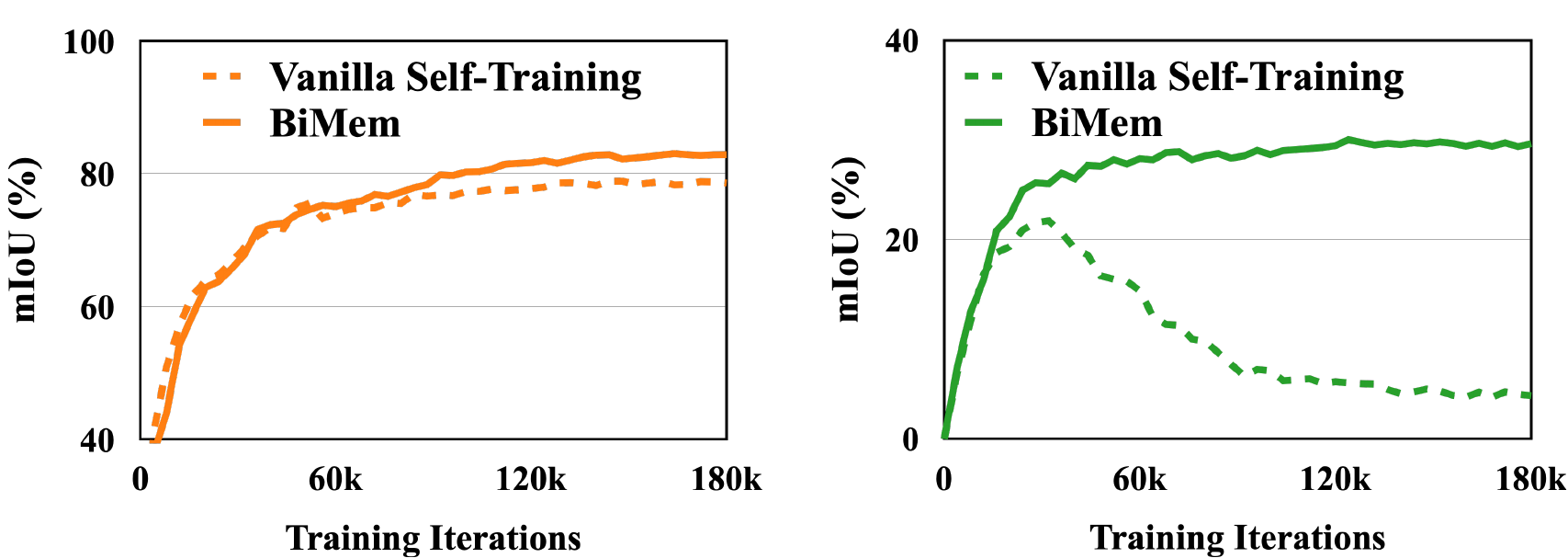}

  \caption{
Analysis of the `forgetting' issue in black-box UDA. We split the target data into two portions according to their initial pseudo-label predictions $Y_t$ by black-box predictor. 
On the target data whose initial pseudo-label predictions are correct in the left graph, the vanilla self-training (trained using all target data) performs consistently well along the training process. 
But on the target data whose initial pseudo-label predictions are incorrect in the right graph, the vanilla self-training performs well at the starting stage and then collapses with training moving on, indicating that the vanilla self-training gradually forgets the earlier learnt useful knowledge. 
As a comparison, the proposed BiMem can remember the learnt useful knowledge and performs consistently well for both portions of target data.
}
\label{fig:dis_forget}
\end{figure}

\noindent \textbf{Analysis of the `Forgetting' Issue in Black-box UDA:}
We examine the source of the `forgetting' illustrated in Fig.~\ref{fig:forget} with a controlled experiment. Specifically, we split the target data into two subsets according to their initial pseudo labels $\hat{Y}_{t}$ (predicted by the black-box predictor). This produces target data with correct initial pseudo label ($i.e.$, $X_{t}^{\text{correct}} = X_{t}[\hat{Y}_{t} = Y_{t}]$) and target data with incorrect initial pseudo labels ($i.e.$, $X_{t}^{\text{incorrect}} = X_{t}[\hat{Y}_{t} \neq Y_{t}]$), where $Y_{t}$ denotes the ground-truth of $X_{t}$ and $X_{t} = X_{t}^{\text{correct}} \cup X_{t}^{\text{incorrect}}$. The splitting allows training models with full data but evaluating them over decomposed data $X_{t}^{\text{correct}}$ and $X_{t}^{\text{incorrect}}$ separately. 
As Fig.~\ref{fig:dis_forget} shows, for vanilla self-training, the mIoU of $X_{t}^{\text{correct}}$ increases stably in the left graph while the mIoU of $X_{t}^{\text{incorrect}}$ increases at the early stage but decreases gradually as shown in the right graph.
This shows that the overall performance degradation at the later training stage mainly comes from $X_{t}^{\text{incorrect}}$, indicating that 
vanilla self-training learns useful information to generate correct predictions for $X_{t}^{\text{incorrect}}$ at the early training stage but tends to forget these information at a later training stage. Differently, BiMem builds comprehensive and robust memorization that memorizes and calibrates useful and representative information on the fly, leading to stabler black-box UDA without performance degradation and training collapse.

We provide more discussions, theoretical insights and qualitative analysis in the appendix.

\section{Conclusion}
This paper presents BiMem, a general black-box UDA framework that works well for various visual recognition tasks.
BiMem constructs three types of memory that interact in a bi-directional manner with a forward memorization flow and a backward calibration flow, resulting in comprehensive yet robust memorization that captures useful and representative information during black-box adaptation.
In this way, BiMem enables to effectively calibrate the noisy pseudo labels conditioned on the memorized information, mitigating the `forgetting' issue in black-box UDA and leading to stable and effective adaptation.
Extensive experiments over multiple benchmarks show that BiMem achieves superior black-box UDA performance consistently across various vision tasks including classification, segmentation, and detection. 
Moving forwards, we plan to further extend our BiMem to other vision tasks such as pose estimation and person re-identification, and investigate other usages of BiMem in addition to label calibration.

\textbf{Acknowledgement.} 
This study is supported under the RIE2020 Industry Alignment Fund – Industry Collaboration Projects (IAF-ICP) Funding Initiative, as well as cash and in-kind contribution from the industry partner(s).

{\small
\bibliographystyle{ieee_fullname}
\bibliography{egbib}
}

\end{document}